\documentclass{article}

\usepackage{microtype}
\usepackage{graphicx}
\usepackage{subfigure}
\usepackage{booktabs} 
\usepackage{dsfont}

\usepackage{times}
\usepackage{xcolor}
\usepackage{soul}
\usepackage[utf8]{inputenc}
\usepackage[small]{caption}
\usepackage[cmex10]{amsmath}
\usepackage{epsfig,amsfonts,amssymb,times,subfigure,floatflt,wrapfig,enumerate,tikz,url,bm}
\usepackage{booktabs}
\usepackage{siunitx}
\usepackage{tabu}
\usepackage{graphicx}  
\usepackage{multirow}
\newcommand{\x}{\mathbf{x}}

\newcommand{\W}[1]{\mathbf{W}^{#1}}

\newcommand{\Au}[1]{\mathbf{\Lambda}^{#1}}
\newcommand{\Al}[1]{\mathbf{\Omega}^{#1}}

\newcommand{\Du}[1]{\mathbf{\lambda}^{#1}}

\newcommand{\upbias}[1]{\mathbf{\Delta}^{#1}}
\newcommand{\lwbias}[1]{\mathbf{\Theta}^{#1}}
\newcommand{\upbiass}[1]{\mathbf{\varphi}^{#1}}

\newcommand{\upbnd}[1]{\mathbf{u}^{#1}}
\newcommand{\lwbnd}[1]{\mathbf{l}^{#1}}

\newcommand{\y}{\mathbf{y}}
\newcommand{\bias}[1]{\mathbf{b}^{#1}}

\newcommand{\upslp}[2]{\mathbf{\alpha}^{#1}_{U,{#2}}}
\newcommand{\lwslp}[2]{\mathbf{\alpha}^{#1}_{L,{#2}}}
\newcommand{\upicp}[2]{\mathbf{\beta}^{#1}_{U,{#2}}}
\newcommand{\lwicp}[2]{\mathbf{\beta}^{#1}_{L,{#2}}}
\newcommand{\upicpp}[2]{\mathbf{\gamma}^{#1}_{U,{#2}}}
\newcommand{\lwicpp}[2]{\mathbf{\gamma}^{#1}_{L,{#2}}}

\allowdisplaybreaks[4]
\usepackage{hyperref}

\usepackage[accepted]{icml2019}

\icmltitlerunning{POPQORN: Quantifying Robustness of Recurrent Neural Networks}

\begin{document}

\twocolumn[
\icmltitle{POPQORN: Quantifying Robustness of Recurrent Neural Networks}

\icmlsetsymbol{equal}{*}

\begin{icmlauthorlist}
\icmlauthor{Ching-Yun Ko}{equal,HKU}
\icmlauthor{Zhaoyang Lyu}{equal,CUHK}
\icmlauthor{Tsui-Wei Weng}{MIT}
\icmlauthor{Luca Daniel}{MIT}
\icmlauthor{Ngai Wong}{HKU}
\icmlauthor{Dahua Lin}{CUHK}
\end{icmlauthorlist}

\icmlaffiliation{HKU}{The University of Hong Kong, Hong Kong}
\icmlaffiliation{CUHK}{The Chinese University of Hong Kong, Hong Kong}
\icmlaffiliation{MIT}{Massachusetts Institute of Technology,
Cambridge, MA, USA. Source code is available at\;\url{https://github.com/ZhaoyangLyu/POPQORN}}

\icmlcorrespondingauthor{Ching-Yun Ko}{cyko@eee.hku.hk}
\icmlcorrespondingauthor{Zhaoyang Lyu}{lyuzhaoyang@link.cuhk.edu.hk}

\icmlkeywords{Robustness, Recurrent neural networks}

\vskip 0.3in
]

\printAffiliationsAndNotice{\icmlEqualContribution} 

\begin{abstract}
The vulnerability to adversarial attacks has been a critical issue for deep neural networks.
Addressing this issue requires a reliable way to evaluate the robustness of a network.
Recently, several methods have been developed to compute \textit{robustness quantification}
for neural networks, namely, certified lower bounds of the minimum adversarial perturbation.
Such methods, however, were devised for feed-forward networks, \textit{e.g.} multi-layer perceptron
or convolutional networks. It remains an open problem to quantify robustness for recurrent networks, especially LSTM and GRU. 
For such networks, there exist additional challenges in computing the robustness quantification,
such as handling the inputs at multiple steps and the interaction between gates and states. 
In this work, we propose \textit{POPQORN}
(\textbf{P}ropagated-\textbf{o}ut\textbf{p}ut \textbf{Q}uantified R\textbf{o}bustness for \textbf{RN}Ns),
a general algorithm to quantify robustness of RNNs, including vanilla RNNs, LSTMs, and GRUs.
We demonstrate its effectiveness on different network architectures and show that the robustness
quantification on individual steps can lead to new insights.

\end{abstract}

\vspace{-1em}
\begin{table*}[tbh!]
\vspace{-1em}
\centering
\caption{Comparison of methods for evaluating RNN robustness.} 
\label{tbl:table_comparison2}
\scalebox{0.75}
{
\begin{tabular}{l|cccccc}
\hline  Method    & Task & Application & Architecture & Attack &  Verification & Robustness guarantee  \\ \hline
FGSM~\cite{papernot2016crafting} & Categorical/ Sequential & NLP & LSTM & \checkmark & $\times$ & $\times$\\
\cite{Yuan2017Crafting} & Categorical & Speech & WaveRNN (RNN/ LSTM) & \checkmark & $\times$ & $\times$ \\
Houdini~\cite{Ciss2017Houdini} & Sequential & Speech & DeepSpeech-2 (LSTM)& \checkmark & $\times$ & $\times$ \\
\cite{Jia2017Adversarial} & SQuAD & NLP & LSTM & \checkmark & $\times$ & $\times$\\ 
\cite{zhao2018generating} & Categorical & NLP & LSTM & \checkmark & $\times$ & $\times$\\
\cite{Ebrahimi2018Adversarial,ebrahimi2018hotflip} & Categorical/ Sequential & NLP & LSTM & \checkmark & $\times$ & $\times$\\
Seq2Sick~\cite{Cheng2018Seq2Sick} & Sequential & NLP & Seq2seq (LSTM) & \checkmark & $\times$ & $\times$\\
C\&W~\cite{Carlini2018Audio} & Sequential & Speech
& DeepSpeech (LSTM) & \checkmark & $\times$ & $\times$\\
CLEVER~\cite{weng2018evaluating} & Categorical & CV/ NLP/ Speech & RNN/ LSTM/ GRU & $\times$ &\checkmark & $\times$   \\\hline \hline
POPQORN (This work) & Categorical & CV/ NLP/ Speech & RNN/ LSTM/ GRU & $\times$ &\checkmark & \checkmark   \\ \hline
\end{tabular}
}

\vspace{-0.2em}

\centering
\caption{Comparison of methods for providing adversarial robustness quantification in NNs.} 
\label{tbl:table_comparison}
\scalebox{0.75}
{
\begin{tabular}{l|cccccccc}
\hline  Method    & Certification & Multi-layer  & Beyond ReLU/ MLP  & RNN structures & handle cross-nonlinearity & Implementation\\ \hline
\cite{hein2017formal} & \checkmark & $\times$ & differentiable/$\times$ &  $\times$ & $\times$ & TensorFlow \\
CLEVER~\cite{weng2018evaluating} & $\times$ & \checkmark & \checkmark/ \checkmark & $\times$ & \checkmark & NumPy\\
SDP approach~\cite{raghunathan2018certified}& \checkmark & $\times$ & \checkmark/ $\times$ &  $\times$  & $\times$ & TensorFlow \\
Dual approach~\cite{dvijotham2018dual}& \checkmark & \checkmark & \checkmark/ \checkmark  & $\times$   &  $\times$ & Not specified \\
Fast-lin~/~Fast-lip \cite{weng2018towards}& \checkmark & \checkmark & $\times$/ $\times$ & $\times$ & $\times$ & NumPy\\
CROWN~\cite{zhang2018crown} & \checkmark & \checkmark & \checkmark/ $\times$ & $\times$ & $\times$ & NumPy\\
DeepZ~\cite{Singh2018Fast} & \checkmark & \checkmark & \checkmark/ no .pooling layers & $\times$ & $\times$ & Python, C\\
CNN-Cert~\cite{Akhilan2019CNN-Cert} & \checkmark & \checkmark & \checkmark/ \checkmark & $\times$ & $\times$ & NumPy\\ \hline \hline
POPQORN (This work) & \checkmark & \checkmark & \checkmark/ \checkmark & \checkmark & \checkmark & PyTorch (GPU)\\ \hline
\end{tabular}
}
\vspace{-0.75em}
\end{table*}

\section{Introduction}
\label{intro}

Deep learning has led to remarkable performance gains on a number of tasks, \textit{i.e.} image classification, speech recognition, and language processing. 
Nevertheless, recent literature has demonstrated that adversarial examples broadly exist for deep neural networks in these applications~\cite{szegedy2014intriguing,kurakin2017adversarial,Jia2017Adversarial,Carlini2018Audio}.  A small perturbation that humans are mostly immune to can be crafted to mislead a neural network's predictions. As deep neural networks have been widely employed in many safety-critical applications~\cite{sharif2016accessorize,kurakin2017adversarial,eykholt2018robust}, it is crucial to know when such models will fail and what can be done to make them more robust to the adversarial attacks. 

Studies on the robustness of neural networks mainly fall in two categories:
(1) \emph{Attack-based approaches} -- researchers try to design strong adversarial attack algorithms to attack deep neural networks and the robustness is measured by the distortion between successful adversarial examples and the original ones. 
(2) \emph{Verification-based approaches}~\cite{katz2017reluplex,cheng2017maximum}, which aim to find the minimum distortion of neural networks or its lower bounds that are \emph{agnostic to attack methods}.
Existing verification-based methods were mainly devised for feed-forward networks, such as 
multi-layer perceptron. Robustness verification for recurrent neural networks (RNNs), which
have been widely used in speech recognition and natural language processing,
has not been systematically investigated.

To verify the robustness of RNNs, we face several new challenges:
(1) Some popular RNN formulations, including LSTM and GRU, involve hidden states that are indirectly determined by three to four gates. These nonlinear gates are tightly coupled together, which we refer to as \textit{cross-nonlinearity}. This substantially complicates the verification of robustness. 
(2) RNNs are widely adopted for applications with sequential inputs, \textit{e.g.}~sentences or time series. However, previous verification methods typically assume that the inputs were fed into the network at the bottom layer. Hence, they are not directly applicable here.
(3) For applications with sequential inputs, imperceptible adversarial examples may correspond to texts with least number of changed words~\cite{Gao2018Black-Box}.
Thus it is critical to evaluate the hardness of manipulating one single word (one input frame) instead of all words. 

In this paper, we tackle the aforementioned problems by proposing an effective robustness quantification framework called POPQORN (\textbf{P}ropagated-\textbf{o}ut\textbf{p}ut \textbf{Q}uantified R\textbf{o}bustness for \textbf{RN}Ns) for RNNs. We bound the non-linear activation function using linear functions. Starting from the output layer, linear bounds are propagated back to the first layer recursively. 
Compared to existing methods, POPQORN has three important advantages:
(1) \textit{Novel} - it is a general framework, which is, to the best of our knowledge, the \textbf{first} work to provide a quantified robustness evaluation for RNNs with robustness guarantees.
(2) \textit{Effective} - it can handle complicated RNN structures besides vanilla RNNs, including LSTMs and GRUs that contain challenging coupled nonlinearities.
(3) \textit{Versatile} - it can be widely applied in applications including but not limited to computer vision, natural language processing, and speech recognition.

\section{Background and Related Works}
\vspace{-0.2em}
\paragraph{Adversarial attacks in RNNs.}
Crafting adversarial examples of RNNs in natural language processing and speech recognition has started to draw public attentions in addition to the adversarial examples of feed-forward networks in image classifications. Adversarial attacks of RNNs on text classification task~\cite{papernot2016crafting}, reading comprehension systems~\cite{Jia2017Adversarial}, seq2seq models~\cite{Cheng2018Seq2Sick} have been proposed in natural language processing application; meanwhile recently~\cite{Yuan2017Crafting} and \cite{Ciss2017Houdini,Carlini2018Audio} have also demonstrated successful attacks in speech recognition and audio systems. We summarize the RNN attacks in Table 1. ~\cite{Papernot2016CraftingAI,Yuan2017Crafting,Ebrahimi2018Adversarial,ebrahimi2018hotflip,Cheng2018Seq2Sick,Carlini2018Audio} perform gradient-based attacks, while generative adversarial network is used in~\cite{zhao2018generating} to generate adversaries. 
We adapt ~\cite{carlini2017towards} into C\&W-Ada for finding RNN attacks.

\vspace{-1em}
\paragraph{Robustness verification for neural networks.} 
To safeguard against misclassification under a threat model, \textit{verification-based} methods evaluate the strengths such that any possible attacks weaker than the proposed strengths will fail. A commonly-used threat model is the norm-ball bounded attacks, wherein strengths of adversaries are quantified by their $l_p$ distance from the original example. Under the norm-ball bounded threat model, determining the minimum adversarial distortion of ReLU networks has been shown to be NP-hard~\cite{katz2017reluplex}. Despite the hardness of the problem, fortunately, it is possible to estimate the minimum adversarial distortion~\cite{weng2018evaluating} or to compute a non-trivial certified lower bound~\cite{hein2017formal,raghunathan2018certified,weng2018towards,dvijotham2018dual,zhang2018crown,Singh2018Fast,Akhilan2019CNN-Cert}. In~\cite{weng2018evaluating}, the authors proposed a robustness score, CLEVER, for neural network image classifiers based on estimating local Lipschitz constant of the networks. We show that it is possible to directly adapt their framework and compute a CLEVER score for RNNs, which is referred to CLEVER-RNN in Section~\ref{sec:Exp}. 

However, CLEVER score does not come with guarantees and therefore an alternative approach is to compute a non-trivial lower bound of the minimum adversarial distortion. \cite{hein2017formal} and~\cite{raghunathan2018certified} proposed to  analytically compute the lower bounds for shallow networks (with no more than $2$ hidden layers) but their bounds get loose easily when applied to general networks.~\cite{weng2018towards} proposed two algorithms Fast-Lin and Fast-Lip to efficiently quantify robustness for ReLU networks and the Fast-Lin algorithm is further extended to CROWN~\cite{zhang2018crown} to quantify for MLP networks with general activation functions. On the other hand,~\cite{dvijotham2018dual} proposed a dual-approach to verify NNs with a general class of activation functions. However, their approach compromises the lower bound performance and the computational efficiency of the ``anytime`` algorithm. Recently, \cite{Akhilan2019CNN-Cert} proposes a framework that is capable of quantifying robustness on CNNs with various architectures including pooling layers, residual blocks in contrast to~\cite{Singh2018Fast} that is limited to convolutional architectures without pooling layers. 

To the best of our knowledge, there is no prior work to provide robustness verification for RNNs that tackles the aforementioned challenges: the complex feed-back architectures, the sequential inputs, and the cross-nonlinearity of the hidden states. In line with the robustness verification for NNs, this paper is the first to study the robustness properties of RNNs (see Tables~\ref{tbl:table_comparison2} and~\ref{tbl:table_comparison} for detailed comparisons). Since it is not straightforward to define a mistake in tasks other than classifications (cf. a mistake in classification tasks refers to a misclassification), we limit the scope of this paper to RNN-based classifiers.

\begin{table*}[t]
\vspace{-1em}
\centering
\caption{Table of Notation}
\scalebox{0.84}
{
\begin{tabular}{ll|ll|ll}
\hline
\textbf{Notation} & \textbf{Definition} & \textbf{Notation} & \textbf{Definition} & \textbf{Notation} & \textbf{Definition}\\
\hline
$t$ & number of output classes & $F: \mathbb{R}^{n\times m} \to \mathbb{R}^{t}$ & network classifier & $\mathbf{X}_0\in\mathbb{R}^{n\times m}$ & original input \\
$n$ & size of input frames & $\x^{(k)}\in\mathbb{R}^n$ & the $k$-th input frame & $[K]$ & set $\{1,2,\cdots,K\}$\\
$s$ & number of neurons in a layer & $F_j^{L}(\mathbf{X}) : \mathbb{R}^{n\times m}\to\mathbb{R}$ & linear lower bound of $F_j(\mathbf{X})$ & \multirow{2}{*}{$\mathbb{B}_p(\x_0^{(k)},\epsilon)$} & \multirow{2}{*}{$\{\x \mid {\|\x-\x_0^{(k)}\|}_p\leq\epsilon\}$}\\
$m$ & number of network layers & $F_j^{U}(\mathbf{X}) : \mathbb{R}^{n\times m}\to\mathbb{R}$ & linear upper bound of $F_j(\mathbf{X})$ & & \\
$\mathbf{a}_0$ & initial hidden state & $\gamma_j^L$ & global lower bound of $F_j(\mathbf{X})$ & \multirow{2}{*}{$\mathbf{l} \preccurlyeq \y \preccurlyeq \mathbf{u}$} & $\mathbf{l}_r \leq \y_r \leq \mathbf{u}_r$,\\
$\mathbf{c}_0$ & initial cell state & $\gamma_j^U$ & global upper bound of $F_j(\mathbf{X})$ & & $\forall\; r\in[s]$, $\mathbf{l}, \y, \mathbf{u}\in\mathbb{R}^s$\\
\hline 
\end{tabular}
}
\label{tbl:notation}
\vspace{-1em}
\end{table*}

\section{POPQORN: Quantifying Robustness of Recurrent Neural Networks}
\vspace{-0.2em}
\paragraph{Overview.} In this section, we show that the output of an RNN can be bounded by two \textit{linear} functions when the input sequence of the network is perturbed within an $\ell_p$ ball with a radius $\epsilon$. By applying the \textit{linear} bounds on the \textit{non-linear} activation functions (e.g. sigmoid and tanh) and the \textit{non-linear} operations (e.g. cell states in LSTM), we can \textit{decouple} the non-linear activations and the \textit{cross-nonlinearity} in the hidden states layer by layer and eventually bound the network output by two \textit{linear} functions in terms of input\footnote{Proposed bounding techniques are applicable to any non-linear activation that is bounded above and below for the given interval.}. Subsequently, we show how this theoretical result is used in the POPQORN to compute robustness quantification of an RNN. For ease of illustrations, we start with two motivating examples on a 2-layer vanilla RNN and a 1-layer LSTM network. The complete theorems for general $m$-layer vanilla RNNs, LSTMs, and GRUs are provided in Section A in the appendix.  

\vspace{-1em}
\paragraph{Notations.}
Let $\mathbf{X_0}=[\x_0^{(1)},\ldots,\x_0^{(m)}]$ be the original input data sequence, and let $\mathbf{X}=[\x^{(1)},\ldots,\x^{(m)}]$ be the perturbed frames of $\mathbf{X_0}$ within an $\epsilon$-bounded $l_p$-ball, i.e., $\x^{(k)}\in\mathbb{B}_p(\x_0^{(k)},\epsilon)$, where superscript ``$(k)$" denotes the $k$-th time step. An RNN function is denoted as $F$ and the $j$-th output element as $F_j(\mathbf{X})$. The upper and lower bounds of $F_j(\mathbf{X})$ are denoted as $F_j^{U}(\mathbf{X})$ and $F_j^{L}(\mathbf{X})$, respectively. The full notations are summarized in Table~\ref{tbl:notation}.

\subsection{A 2-layer Vanilla RNN}
\label{Exm:1}
\vspace{-0.2em}

\begin{figure}[t]
    \centering
    \includegraphics[width=0.4\textwidth]{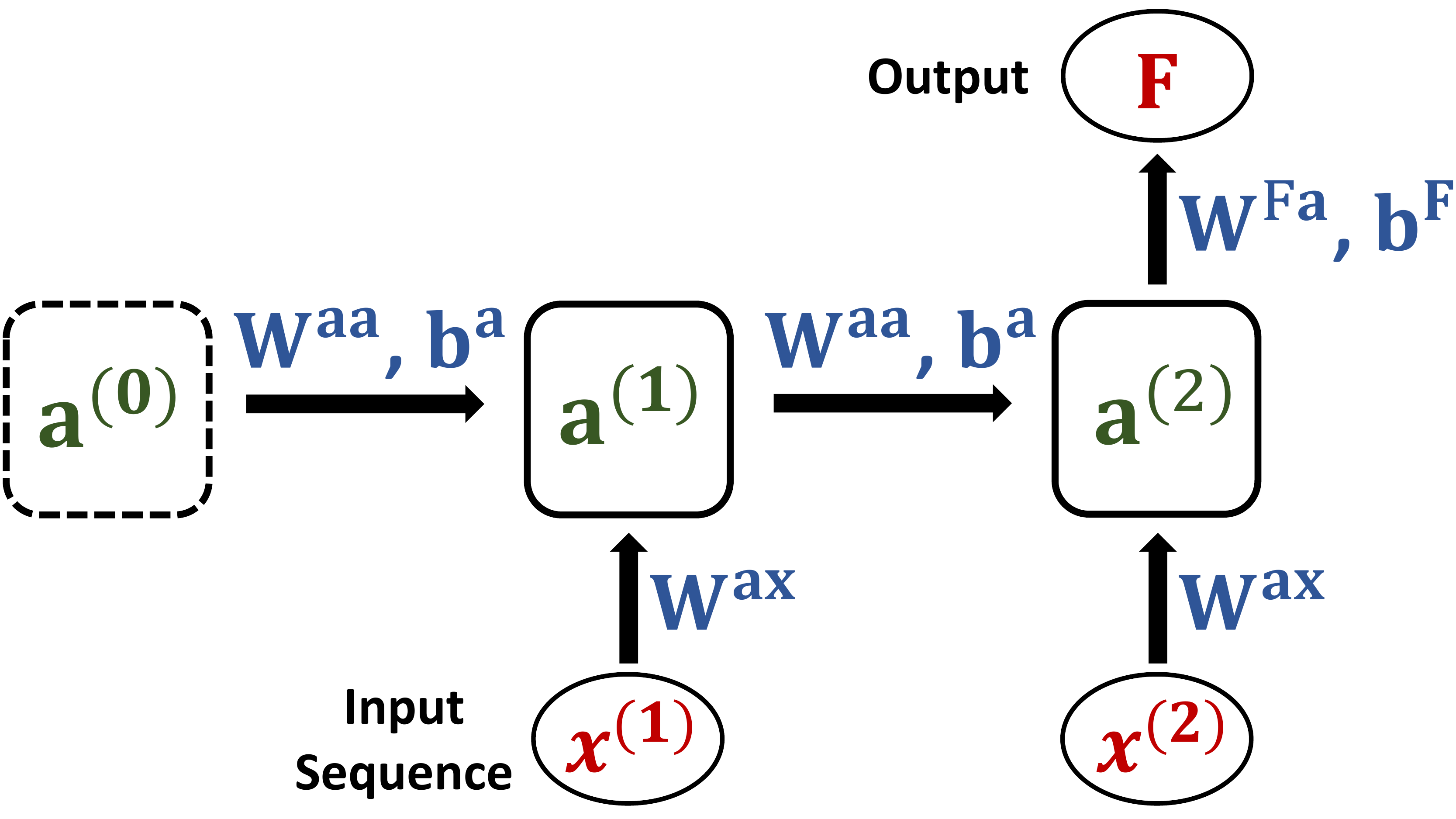}
    \caption{Graphical depiction of a $2$-layer many-to-one RNN.}
    \label{fig:2LRNN}
    \vspace{-1em}
\end{figure}

\paragraph{Definition.}
A many-to-one $2$-layer ($m=2$) RNN reads
\begin{align*}
   F(\mathbf{X}) &= \W{Fa} \mathbf{a}^{(2)}+\mathbf{b}^{F},\\
   \mathbf{a}^{(2)} &= \sigma(\W{aa} \mathbf{a}^{(1)}+\W{ax} \x^{(2)}+\mathbf{b}^{a}),\\
   \mathbf{a}^{(1)} &= \sigma(\W{aa} \mathbf{a}^{(0)}+\W{ax} \x^{(1)}+\mathbf{b}^{a}),
\end{align*}
where $F$ is the output, $\mathbf{a}^{(k)}$ is the $k$-th hidden state, $\sigma(\cdot)$ is the coordinate-wise activation function, and $\W{aa},\W{Fa},\mathbf{b}^{F},\mathbf{b}^{a}$ are associated model parameters, as shown in Figure~\ref{fig:2LRNN}.

\vspace{-0.5em}
\paragraph{Ideas.} Based on the above equation, for a fixed $j\in[t]$, we aim at deriving two explicit functions $F_j^{L}$ and $F_j^{U}$ such that $\forall~\mathbf{X}\in\mathbb{R}^{n\times 2}$ where $\x^{(k)}\in\mathbb{B}_p(\x_0^{(k)},\epsilon)$, the inequality $F_j^{L}(\mathbf{X})\leq F_j(\mathbf{X})\leq F_j^{U}(\mathbf{X})$ holds true. To start with, we assume that we know the bounds for \textit{pre-activation} $\y^{(k)}=\W{aa} \mathbf{a}^{(k-1)}+\W{ax} \x^{(k)}+\mathbf{b}^{a}$ (superscript ``$(k)$" denotes the $k$-th layer): 
$\lwbnd{(k)} \preccurlyeq \y^{(k)} \preccurlyeq \upbnd{(k)}.$ We refer the $\mathbf{l}$ and $\mathbf{u}$ as \textit{pre-activation bounds} in the remainder of this paper. We show that it is possible to bound every non-linear activation function $\sigma(\cdot)$ in the hidden states by two bounding lines in  \eqref{eq:def1a} and \eqref{eq:def1b} associated to $\lwbnd{(k)}$ and $\upbnd{(k)}$, and that we can apply this procedure recursively from the output $F_j(\mathbf{X})$ to the input $\mathbf{X}$ to obtain $F_j^{L}$ and $F_j^{U}$.

\vspace{-0.5em}
\paragraph{Bounding lines.} Two univariate bounding linear functions are defined as follows:
\begin{subequations}
\begin{align}
\label{eq:def1a}
    h^{(k)}_{U,r}(\mathbf{v})&=\upslp{(k)}{r}(\mathbf{v}+\upicp{(k)}{r}),\\
\label{eq:def1b}
    h^{(k)}_{L,r}(\mathbf{v})&=\lwslp{(k)}{r}(\mathbf{v}+\lwicp{(k)}{r}), 
\end{align}
\end{subequations}such that for $\lwbnd{(k)}_r\leq \mathbf{v}\leq \upbnd{(k)}_r$,
\begin{align*}
    \text{eq.} \eqref{eq:def1b} \leq \sigma(\mathbf{v}) \leq \text{eq.} \eqref{eq:def1a},
\end{align*}
where the $r$ in the subscript implies the dependency of the derived lines on neurons. Both the slopes and intercepts are functions of \textit{pre-activation} bounds.

\vspace{-0.5em}
\paragraph{Derivation.}
We exemplify how a $2$-layer vanilla RNN can be bounded:
\begin{align}
    F_j(\mathbf{X}) &= \W{Fa}_{j,:} \mathbf{a}^{(2)}+\bias{F}_j= \W{Fa}_{j,:} \sigma(\y^{(2)})+\bias{F}_j.
    \label{eqn:exm1.1}
\end{align}
We use $s$ upper-bounding lines $h^{(2)}_{U,r}(\y^{(2)}_r), r\in[s]$, and also use variables $\Du{(2)}_{j,r}$ and $\upbias{(2)}_{r,j}$ to denote the slopes in front of $\y^{(2)}_r$ and intercepts in the parentheses:
\begingroup
\small
\begin{align*}
    \Du{(2)}_{j,r}  &=
    \begin{cases}
    \upslp{(2)}{r}\;\;\text{if}\;\; \W{Fa}_{j,r}\geq 0;\\
    \lwslp{(2)}{r}\;\;\text{if}\;\; \W{Fa}_{j,r}< 0;
    \end{cases}\!\!\!\!
    \upbias{(2)}_{r,j}  =
    \begin{cases}
    \upicp{(2)}{r}\;\;\text{if}\;\; \W{Fa}_{j,r}\geq 0;\\
    \lwicp{(2)}{r}\;\;\text{if}\;\; \W{Fa}_{j,r}< 0,
    \end{cases}
\end{align*}
\endgroup
and obtain
\begin{align*}
    F_j(\mathbf{X}) &\leq (\W{Fa}_{j,:}\; \odot\Du{(2)}_{j,:})(\y^{(2)}+\upbias{(2)}_{:,j})+\bias{F}_j,
\end{align*}
where $\odot$ is the Hadamard (i.e. element-wise) product. To simplify notation, we let $\Au{(2)}_{j,:} := \W{Fa}_{j,:}\; \odot\Du{(2)}_{j,:}$ and have
\begin{align*}
    F_j(\mathbf{X})&\leq 
    \tilde{\mathbf{W}}^{aa(2)}_{j,:} \mathbf{a}^{(1)}+\tilde{\mathbf{W}}^{ax(2)}_{j,:} \x^{(2)}+\tilde{\mathbf{b}}^{(2)}_j,
\end{align*}
where
\begin{align*}
    \tilde{\mathbf{W}}^{aa(2)}_{j,:} \nonumber&= \Au{(2)}_{j,:}\W{aa},\ 
    \tilde{\mathbf{W}}^{ax(2)}_{j,:} = \Au{(2)}_{j,:}\W{ax},\\
    \tilde{\mathbf{b}}^{(2)}_j \nonumber&= \Au{(2)}_{j,:}(\bias{a}+\upbias{(2)}_{:,j})+\bias{F}_j.
\end{align*}
Substituting $\mathbf{a}^{(1)}$ with its definition yields
\begin{align}
    F_j(\mathbf{X}) &\leq \tilde{\mathbf{W}}^{aa(2)}_{j,:}\sigma(\y^{(1)})+\tilde{\mathbf{W}}^{ax(2)}_{j,:} \x^{(2)}+\tilde{\mathbf{b}}^{(2)}_j.
    \label{eqn:exm1.2}
\end{align}
Note that Equations~\eqref{eqn:exm1.1} and~\eqref{eqn:exm1.2} are in similar forms. Thus we can bound $\sigma(\y^{(1)})$ by $s$ linear functions $h^{(1)}_{U,r}(\y^{(1)}_r), r\in[s]$ and use $\Du{(1)}_{j,r}$ and $\upbias{(1)}_{r,j}$ to denote slopes in front of $\y^{(1)}_r$ and intercepts in the parentheses:
\begin{align}
    \Du{(1)}_{j,r} \nonumber& =
    \begin{cases}
    \upslp{(1)}{r}\;\;\;\text{if}\;\;\; \tilde{\mathbf{W}}^{aa(2)}_{j,r}\geq 0;\\
    \lwslp{(1)}{r}\;\;\;\text{if}\;\;\; \tilde{\mathbf{W}}^{aa(2)}_{j,r}< 0;
    \end{cases}\\
    \upbias{(1)}_{r,j} \nonumber& =
    \begin{cases}
    \upicp{(1)}{r}\;\;\;\text{if}\;\;\; \tilde{\mathbf{W}}^{aa(2)}_{j,r}\geq 0;\\
    \lwicp{(1)}{r}\;\;\;\text{if}\;\;\; \tilde{\mathbf{W}}^{aa(2)}_{j,r}< 0,
    \end{cases}
\end{align}
and let $\Au{(1)}_{j,:} := \tilde{\mathbf{W}}^{aa(2)}_{j,:}\odot \Du{(1)}_{j,:}$. Then we have
\begin{align*}
    F_j(\mathbf{X})
    &\leq\Au{(1)}_{j,:}\W{aa} \mathbf{a}^{(0)}+\sum^{2}_{z=1}\Au{(z)}_{j,:}\W{ax} \x^{(z)}\\
    &+\sum^{2}_{z=1}\Au{(z)}_{j,:}(\bias{a}+\upbias{(z)}_{:,j})+\bias{F}_j.
\end{align*}
So far we have derived an explicit linear function on the right-hand side of the above inequality. We denote it herein as $F^U_j$ and we have $F_j(\mathbf{X})\leq F^U_j(\mathbf{X})$, $\forall~\mathbf{X}\in\mathbb{R}^{n\times 2}$ where $\x^{(k)}\in\mathbb{B}_p(\x_0^{(k)},\epsilon)$. A closed-form global upper bound $\gamma^U_j$ can be obtained naturally by maximizing $F^U_j(\mathbf{X})$ for all possible $\mathbf{X}$, which can be directly solved by applying Holder's inequality. Hence we obtain
\begin{align*}
    \gamma^U_j&=\Au{(1)}_{j,:}\W{aa} \mathbf{a}^{(0)}+\sum^{2}_{z=1}\epsilon\|\Au{(z)}_{j,:}\W{ax}\|_q\\
    &+\sum^{2}_{z=1}\Au{(z)}_{j,:}\W{ax} \x^{(z)}_0+\sum^{2}_{z=1}\Au{(z)}_{j,:}(\bias{a}+\upbias{(z)}_{:,j})+\bias{F}_j.
\end{align*}
An explicit function $F^L_j$ and a closed-form global lower bound $\gamma^L_j$ can also be found through similar steps and by minimizing $F^L_j(\mathbf{X})$ instead. During the derivation, we have limited perturbations to be uniform across input frames, which assemble noises in cameras. However, the above method is also applicable to non-uniform distortions. For example, we can also certify bounds for distortions on parts of the input frames, which on the other hand are of more interests in natural language processing tasks.

\subsection{A 1-layer Long Short-term Memory Network}
\label{Exm:2}
\begin{figure}[t]
    \centering
    \includegraphics[width=0.45\textwidth]{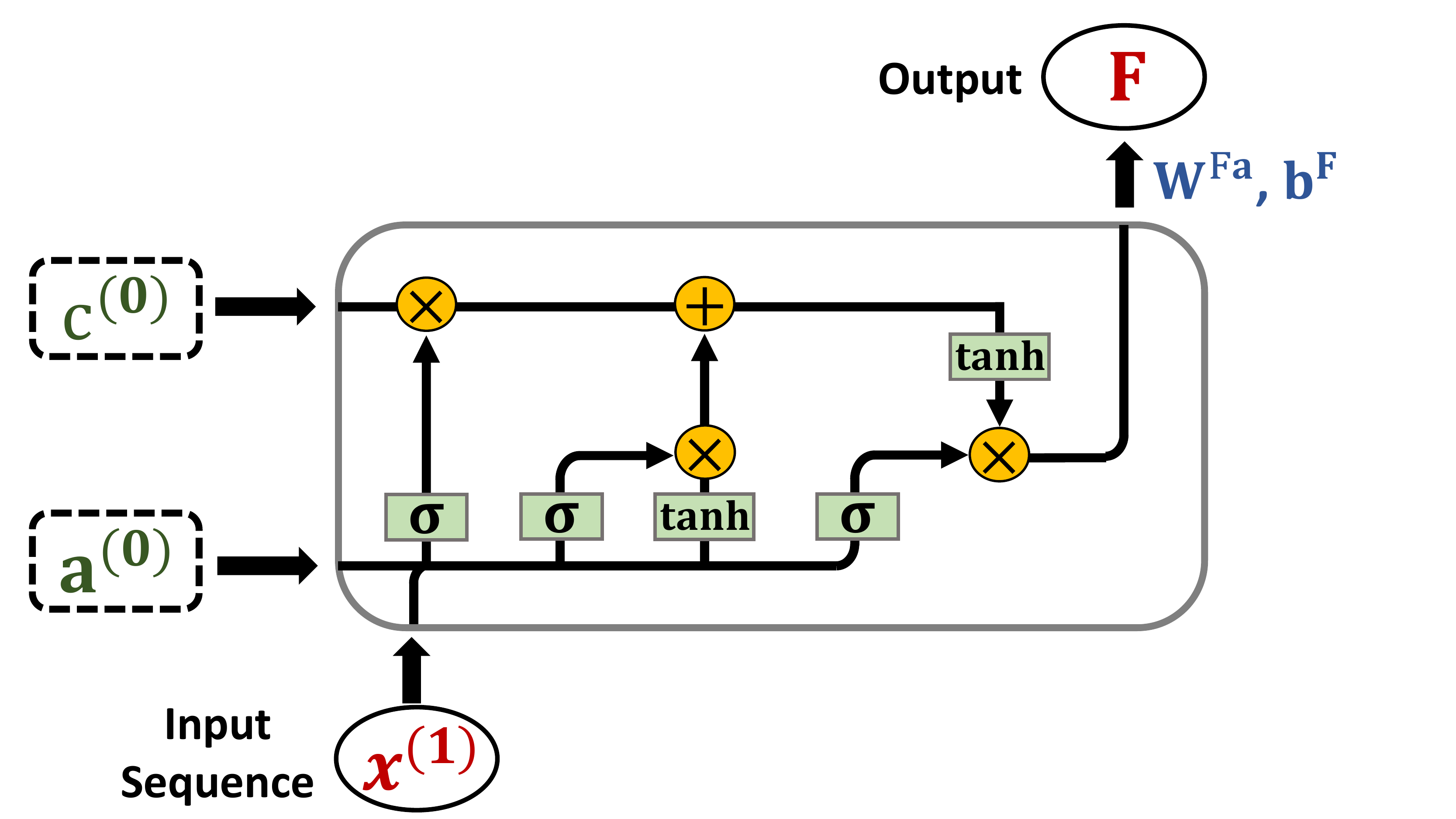}
    \caption{Graphical depiction of a $1$-layer LSTM network}
    \label{fig:1LLSTM}
    \vspace{-1em}
\end{figure}

For an RNN with LSTM units, we can also derive analytic upper-bounding and lower-bounding functions. 
In this example, we inherit the notations in Table~\ref{tbl:notation}, and exemplify through a $1$-layer ($m=1$) LSTM as shown in Figure~\ref{fig:1LLSTM}. 

\vspace{-0.5em}
\paragraph{LSTM.} The following updating equations are adopted:
\begin{align*}
    \text{Input gate:} \; \mathbf{i}^{(k)}
    &=\sigma(\W{ix}\x^{(k)}+\W{ia}\mathbf{a}^{(k-1)}+\bias{i});\\
    \text{Forget gate:} \; \mathbf{f}^{(k)}
    &=\sigma(\W{fx}\x^{(k)}+\W{fa}\mathbf{a}^{(k-1)}+\bias{f});\\
    \text{Cell gate:} \; \mathbf{g}^{(k)}
    &=\tanh(\W{gx}\x^{(k)}+\W{ga}\mathbf{a}^{(k-1)}+\bias{g});\\
    \text{Output gate:} \; \mathbf{o}^{(k)}
    &=\sigma(\W{ox}\x^{(k)}+\W{oa}\mathbf{a}^{(k-1)}+\bias{o});\\
    \text{Cell state:} \; \mathbf{c}^{(k)}&= \mathbf{f}^{(k)}\odot \mathbf{c}^{(k-1)}+\mathbf{i}^{(k)}\odot \mathbf{g}^{(k)};\\
    \text{Hidden state:} \; \mathbf{a}^{(k)}&= \mathbf{o}^{(k)}\odot \tanh(\mathbf{c}^{(k)}).
\end{align*}
Again, the $\sigma(\cdot)$ denotes the coordinate-wise sigmoid function and $\tanh(\cdot)$ is the coordinate-wise hyperbolic tangent function. Output $F$ of the network is determined by $F(\mathbf{X}) = \W{Fa} \mathbf{a}^{(m)}+\mathbf{b}^{F}$, and we have pre-activations as
\begin{align*}
    \y^{i(k)} &= \W{ix}\x^{(k)}+\W{ia}\mathbf{a}^{(k-1)}+\bias{i};\\
    \y^{f(k)} &= \W{fx}\x^{(k)}+\W{fa}\mathbf{a}^{(k-1)}+\bias{f};\\
    \y^{g(k)} &= \W{gx}\x^{(k)}+\W{ga}\mathbf{a}^{(k-1)}+\bias{g};\\
    \y^{o(k)} &= \W{ox}\x^{(k)}+\W{oa}\mathbf{a}^{(k-1)}+\bias{o}.
\end{align*}
\vspace{-2em}
\paragraph{Ideas.}
When deriving the lower bounds and upper bounds of LSTMs, we have also made an important assumption: we assume we know the bounds for hidden/cell states. This assumption, as well as the assumption made when deriving for vanilla RNNs and GRUs, can be easily fulfilled as the bounds of hidden/cell states are available by similar derivations as will be shown below. We refer readers to Theorem A.2 (vanilla RNNs), Corollary A.3/ Theorem A.4 (LSTMs) and Corollary A.6 (GRUs) in the appendix for details.

\begin{figure}[t]
    \centering
    \includegraphics[width=0.48\textwidth]{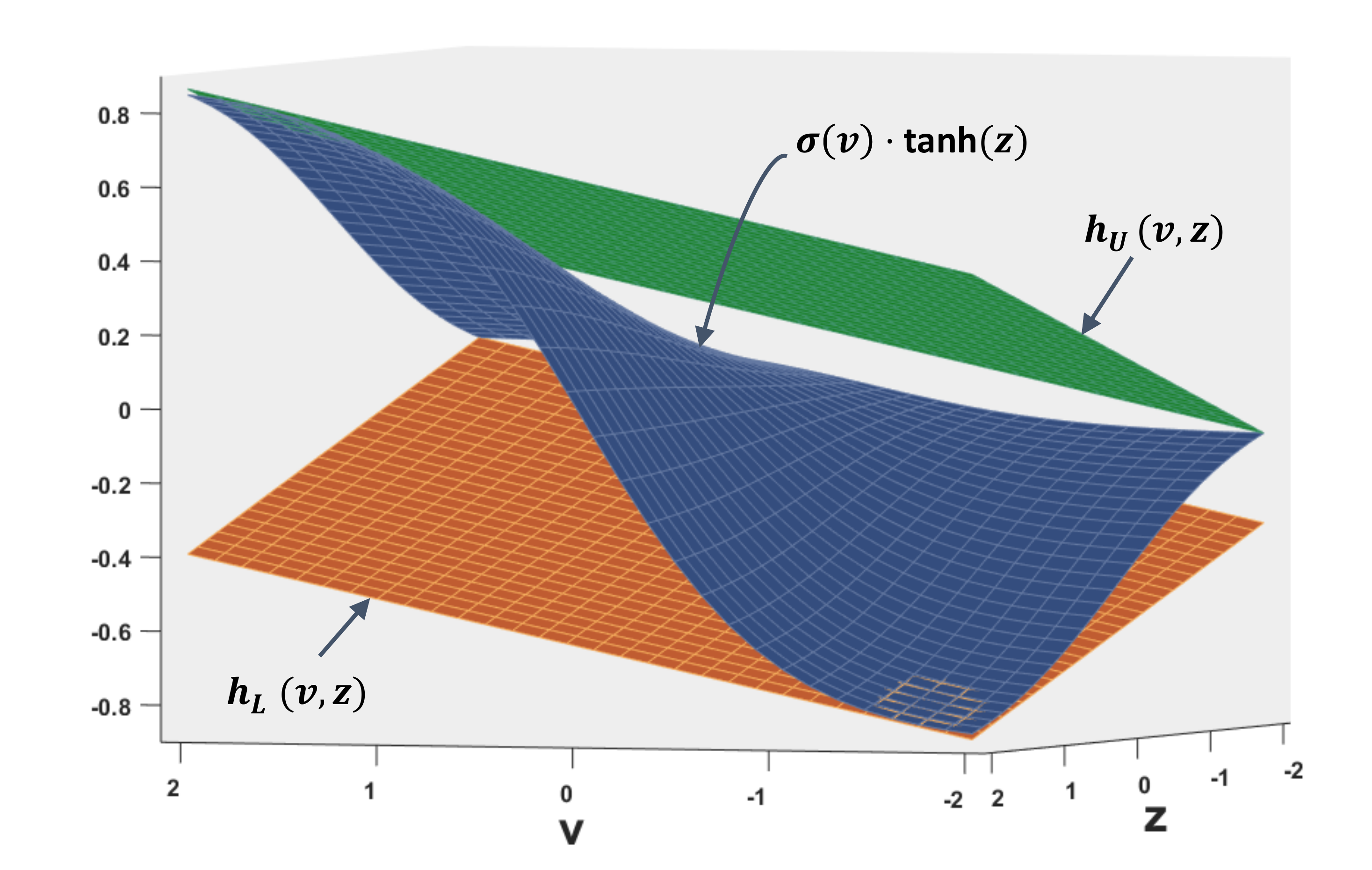}
    \caption{Illustration of the upper-bounding plane $h_U(\mathbf{v},\mathbf{z})$ and lower-bounding plane $h_L(\mathbf{v},\mathbf{z})$ of $\sigma(\mathbf{v})  \tanh(\mathbf{z})$.}
    \label{fig:bdplanes}
    \vspace{-1em}
\end{figure}

Recalling that in Section~\ref{Exm:1}, we aim at bounding the non-linearity using univariate linear functions. Final bounds are obtained by recursively propagating the linear bounds from the output layer back to the first layer. Here when analyzing the bounds for an LSTM, we adopt a similar approach of propagating linear bounds from the last layer to the first. However, different from the vanilla RNN, the difficulty of reaching this goal lies in bounding more complex non-linearities. In an LSTM, we cope with two different non-linear functions: 
\vspace{-0.5em}$$\sigma(\mathbf{v})  \mathbf{z} \quad\text{and}\quad \sigma(\mathbf{v})  \tanh(\mathbf{z}),\vspace{-0.25em}$$
both of which are dependent on two variables and are cross terms of varied gates. To deal with this, we extend our previous ideas to using \textit{planes} instead of \textit{lines} to bound these cross terms. The graphical illustration of the bounding planes is shown in Figure~\ref{fig:bdplanes}.

\vspace{-0.5em}
\paragraph{Bounding planes.}
With bounds of hidden/cell states, we can gather bounds: $\lwbnd{\text{gate}(1)} \preccurlyeq \y^{\text{gate}(1)} \preccurlyeq \upbnd{\text{gate}(1)}$ where gate = $\{i,f,g,o\}$ and superscript ``$(1)$" denotes the layer we are at. As both variables are bounded as above and compactness is a continuous invariant, we are guaranteed to have at least two bounding planes for the non-linearities. For example, for $\sigma(\mathbf{v})  \mathbf{z}$ we have planes: $$h^{\text{cross}(1)}_{U,r}=\max(\sigma(\mathbf{v})  \mathbf{z}),\quad h^{\text{cross}(1)}_{L,r}=\min(\sigma(\mathbf{v})  \mathbf{z}).$$  In practice, we use the following planes:
\begin{subequations}
\begin{align}
\label{eq:def2a}
    h^{\text{cross}(1)}_{U,r}(\mathbf{v},\mathbf{z}) &=\alpha^{\text{cross}(1)}_{U,r}\mathbf{v}+\beta^{\text{cross}(1)}_{U,r} \mathbf{z}+ \gamma^{\text{cross}(1)}_{U,r}, \\
\label{eq:def2b}
    h^{\text{cross}(1)}_{L,r}(\mathbf{v},\mathbf{z}) &=\alpha^{\text{cross}(1)}_{L,r}\mathbf{v}+\beta^{\text{cross}(1)}_{L,r} \mathbf{z}+ \gamma^{\text{cross}(1)}_{L,r},
\end{align}
\end{subequations}that satisfy
\begin{equation*}
    \text{eq.} \eqref{eq:def2b} \leq \sigma(\mathbf{v})  \mathbf{z}\leq \text{eq.} \eqref{eq:def2a},
\end{equation*}
where the $r$ in the subscript implies the dependency of the derived planes on neurons, and superscript cross$\in \{ig,oc,fc\}$ tracks the origins of those cross terms. For example, \mbox{cross$=ig$} when it is the coupling of input gates and cell gates: $\mathbf{v}=\y^{i(1)}$ and $\mathbf{z}=\y^{g(1)}$.
Notably, slopes and intercepts, say $\alpha^{ig(1)}_{U,r}$ depends on ranges of $\y^{i(1)}_r$ and $\y^{g(1)}_r$. We formulate the task of finding bounding planes as a constrained optimization problem and use gradient descent method to solve it (see Section A.4 in the appendix for details).

\begin{table*}[tbh!]
\vspace{-1em}
\centering
\caption{Quantified robustness bounds for various RNNs.} 
\label{tbl:bounds}
\scalebox{0.85}
{
\begin{tabular}{l|c|c}
\hline  Networks    & $\gamma_j^L\leq F_j\leq \gamma_j^U$  & Closed-form formulas \\ \hline
\multirow{2}{*}{Vanilla RNN} & Upper bounds $\gamma_j^U$ & $\Au{(0)}_{j,:}\mathbf{a}^{(0)}+\sum^{m}_{k=1}\epsilon\|\Au{(k)}_{j,:}\W{ax}\|_q+\sum^{m}_{k=1}\Au{(k)}_{j,:}\W{ax}\x_0^{(k)}+\sum^{m}_{k=1}\Au{(k)}_{j,:}(\bias{a}+\upbias{(k)}_{:,j})+\bias{F}_j$ \\
& Lower bound $\gamma_j^L$ & $\Al{(0)}_{j,:}\mathbf{a}^{(0)}-\sum^{m}_{k=1}\epsilon\|\Al{(k)}_{j,:}\W{ax}\|_q+\sum^{m}_{k=1}\Al{(k)}_{j,:}\W{ax}\x_0^{(k)}+\sum^{m}_{k=1}\Al{(k)}_{j,:}(\bias{a}+\lwbias{(k)}_{:,j})+\bias{F}_j$\\ \hline
\multirow{2}{*}{LSTM} & Upper bounds $\gamma_j^U$ & $\mathbf{\tilde{W}}^{a(1)}_{U,j,:}\mathbf{a}^{(0)}+\Au{fc(1)}_{\Delta,j,:}\mathbf{c}^{(0)}+\sum^{m}_{k=1}\epsilon\|\mathbf{\tilde{W}}^{x(k)}_{U,j,:}\|_q+\sum^{m}_{k=1}\mathbf{\tilde{W}}^{x(k)}_{U,j,:}\x^{(k)}_0+\sum^{m}_{k=1}\mathbf{\tilde{b}}^{(k)}_{U,j}+\bias{F}_j$  \\
 & Lower bound $\gamma_j^L$ &  $\mathbf{\tilde{W}}^{a(1)}_{L,j,:}\mathbf{a}^{(0)}+\Al{fc(1)}_{\Theta,j,:}\mathbf{c}^{(0)}-\sum^{m}_{k=1}\epsilon\|\mathbf{\tilde{W}}^{x(k)}_{L,j,:}\|_q+\sum^{m}_{k=1}\mathbf{\tilde{W}}^{x(k)}_{L,j,:}\x^{(k)}_0+\sum^{m}_{k=1}\mathbf{\tilde{b}}^{(k)}_{L,j}+\bias{F}_j$ \\ \hline
\multirow{2}{*}{GRU} & Upper bounds $\gamma_j^U$ & $\mathbf{\tilde{W}}^{a(1)}_{U,j,:}\mathbf{a}^{(0)}+\sum^{m}_{k=1}\epsilon\|\mathbf{\tilde{W}}^{x(k)}_{U,j,:}\|_q+\sum^{m}_{k=1}\mathbf{\tilde{W}}^{x(k)}_{U,j,:}\x^{(k)}_0+\sum^{m}_{k=1}\mathbf{\tilde{b}}^{(k)}_{U,j}+\bias{F}_j$  \\
 & Lower bound $\gamma_j^L$ & $\mathbf{\tilde{W}}^{a(1)}_{L,j,:}\mathbf{a}^{(0)}-\sum^{m}_{k=1}\epsilon\|\mathbf{\tilde{W}}^{x(k)}_{L,j,:}\|_q+\sum^{m}_{k=1}\mathbf{\tilde{W}}^{x(k)}_{L,j,:}\x^{(k)}_0+\sum^{m}_{k=1}\mathbf{\tilde{b}}^{(k)}_{L,j}+\bias{F}_j$  \\ \hline
\multicolumn{3}{l}{Remark: see Section A in the appendix for detailed proofs and definitions.} \\ \hline
\end{tabular}
}
\vspace{-1em}
\end{table*}

\vspace{-0.5em}
\paragraph{Derivation.}
We now exemplify how a $1$-layer LSTM can be bounded:
\begin{align}
    F_j(\mathbf{X}) &= \W{Fa}_{j,:} \mathbf{a}^{(1)}+\bias{F}_j,\nonumber\\
    &= \W{Fa}_{j,:} [\sigma(\y^{o(1)})\odot\tanh(\mathbf{c}^{(1)})]+\bias{F}_j.\label{eqn:lstmFjX}
\end{align}
To bound Equation~(\ref{eqn:lstmFjX}), we use $s$ upper-bounding planes $h^{oc(1)}_{U,r}(\y^{o(1)},\mathbf{c}^{(1)}), r\in[s]$,
and also define $\Du{oc(1)}_{j,r}$, $\upbias{oc(1)}_{j,r}$ and $\upbiass{oc(1)}_{j,r}$ in the parentheses:
\begin{align*}
    \Du{oc(1)}_{j,r}  =
    \begin{cases}
    \upslp{oc(1)}{r}\;\;\text{if}\;\; \W{Fa}_{j,r}\geq 0;\\
    \lwslp{oc(1)}{r}\;\;\text{if}\;\; \W{Fa}_{j,r}< 0;
    \end{cases}\\
    \upbias{oc(1)}_{j,r}  =
    \begin{cases}
    \upicp{oc(1)}{r}\;\;\text{if}\;\; \W{Fa}_{j,r}\geq 0;\\
    \lwicp{oc(1)}{r}\;\;\text{if}\;\; \W{Fa}_{j,r}< 0;
    \end{cases}\\
    \upbiass{oc(1)}_{j,r}  =
    \begin{cases}
    \upicpp{oc(1)}{r}\;\;\text{if}\;\; \W{Fa}_{j,r}\geq 0;\\
    \lwicpp{oc(1)}{r}\;\;\text{if}\;\; \W{Fa}_{j,r}< 0;
    \end{cases}
\end{align*}
and obtain
\begin{align*}
    F_j(\mathbf{X}) &\leq (\W{Fa}_{j,:}\odot\Du{oc(1)}_{j,:})\y^{o(1)}+(\W{Fa}_{j,:}\odot\upbias{oc(1)}_{j,:})\mathbf{c}^{(1)}\\
    &+\sum_{r=1}^s(\W{Fa}_{j,r}\upbiass{oc(1)}_{j,r})+\bias{F}_j.
\end{align*}
For simplicity, we further collect the summing weights of $\y^{o(1)}$ and $\mathbf{c}^{(1)}$ into row vectors $\Au{oc(1)}_{\lambda,j,:}$ and $\Au{oc(1)}_{\Delta,j,:}$, and constants (excepts $\bias{F}_j$) into $\Au{oc(1)}_{\varphi,j,r}$. Thereafter, we have
\begin{align}
    &F_j(\mathbf{X})\nonumber\\
    &\leq (\Au{oc(1)}_{\lambda,j,:}\W{ox})\x^{(1)}+(\Au{oc(1)}_{\lambda,j,:}\W{oa})\mathbf{a}^{(0)}+\Au{oc(1)}_{\lambda,j,:}\bias{o}\nonumber\\
    &+\Au{oc(1)}_{\Delta,j,:}[(\sigma(\y^{f(1)})\odot \mathbf{c}^{(0)}+\sigma(\y^{i(1)})\odot\tanh(\y^{g(1)})]\nonumber\\
    &+\sum^s_{r=1}\Au{oc(1)}_{\varphi,j,r}+\bias{F}_j.\label{eqn:FjXineq}
\end{align}
Then to bound the two cross terms in Equation~(\ref{eqn:FjXineq}), we use upper-bounding planes $h^{fc(1)}_{U,r}(\y^{f(1)},\mathbf{c}^{(0)})$ and $h^{ig(1)}_{U,r}(\y^{i(1)},\y^{g(1)}), r\in[s]$. We define $\Du{\text{cross}'(1)}_{j,r}$, $\upbias{\text{cross}'(1)}_{j,r}$, $\upbiass{\text{cross}'(1)}_{j,r}$ for cross$'$ $=\{fc,ig\}$ as in the parentheses:
\begin{align*}
    \Du{\text{cross}'(1)}_{j,r} & =
    \begin{cases}
    \upslp{\text{cross}'(1)}{r}\;\;\;\text{if}\;\;\; \Au{oc(1)}_{\Delta,j,r}\geq 0;\\
    \lwslp{\text{cross}'(1)}{r}\;\;\;\text{if}\;\;\; \Au{oc(1)}_{\Delta,j,r}< 0;
    \end{cases}\\
    \upbias{\text{cross}'(1)}_{j,r} & =
    \begin{cases}
    \upicp{\text{cross}'(1)}{r}\;\;\;\text{if}\;\;\; \Au{oc(1)}_{\Delta,j,r}\geq 0;\\
    \lwicp{\text{cross}'(1)}{r}\;\;\;\text{if}\;\;\; \Au{oc(1)}_{\Delta,j,r}< 0;
    \end{cases}\\
    \upbiass{\text{cross}'(1)}_{j,r} & =
    \begin{cases}
    \upicpp{\text{cross}'(1)}{r}\;\;\;\text{if}\;\;\; \Au{oc(1)}_{\Delta,j,r}\geq 0;\\
    \lwicpp{\text{cross}'(1)}{r}\;\;\;\text{if}\;\;\; \Au{oc(1)}_{\Delta,j,r}< 0;
    \end{cases}
\end{align*}
and obtain 
\begin{align*}
    F_j(\mathbf{X}) &\leq \mathbf{p}_x\x^{(1)}+\mathbf{p}_a\mathbf{a}^{(0)}+\Au{fc(1)}_{\lambda,j,:}\y^{f(1)}+\Au{fc(1)}_{\Delta,j,:}\mathbf{c}^{(0)}\\
    &+\Au{ig(1)}_{\lambda,j,:}\y^{i(1)}+\Au{ig(1)}_{\Delta,j,:}\y^{g(1)}+\mathbf{p}_b+\bias{F}_j.
\end{align*}
where
\begin{align*}
    \mathbf{p}_x&= \Au{oc(1)}_{\lambda,j,:}\W{ox},\qquad\qquad\, \mathbf{p}_a= \Au{oc(1)}_{\lambda,j,:}\W{oa},\\
    \Au{fc(1)}_{\lambda,j,:}&= \Au{oc(1)}_{\Delta,j,:}\odot\Du{fc(1)}_{j,:},\quad
    \Au{fc(1)}_{\Delta,j,:}= \Au{oc(1)}_{\Delta,j,:}\odot\upbias{fc(1)}_{j,:},\\
    \Au{ig(1)}_{\lambda,j,:}&= \Au{oc(1)}_{\Delta,j,:}\odot\Du{ig(1)}_{j,:},\quad\;
    \Au{ig(1)}_{\Delta,j,:}= \Au{oc(1)}_{\Delta,j,:}\odot\upbias{ig(1)}_{j,:},\\
    \Au{fc(1)}_{\varphi,j,:}&=\Au{oc(1)}_{\Delta,j,:}\odot\upbiass{fc(1)}_{j,:},\quad\, \Au{ig(1)}_{\varphi,j,:}= \Au{oc(1)}_{\Delta,j,:}\odot\upbiass{ig(1)}_{j,:},\\
    \mathbf{p}_b&= \sum^s_{r=1}(\Au{fc(1)}_{\varphi,j,r}+\Au{ig(1)}_{\varphi,j,r}+\Au{oc(1)}_{\varphi,j,r})+\Au{oc(1)}_{\lambda,j,:}\bias{o}.
\end{align*}
Substituting $\y^{f(1)},\mathbf{c}^{(0)},\y^{i(1)},\y^{g(1)}$ with their definitions renders
\begin{align*}
     F_j(\mathbf{X}) &\leq \mathbf{\tilde{W}}^{x(1)}_{U,j,:}\x^{(1)} +\mathbf{\tilde{W}}^{a(1)}_{U,j,:}\mathbf{a}^{(0)}+\mathbf{\tilde{b}}^{(1)}_{U,j}+\Au{fc(1)}_{\Delta,j,:}\mathbf{c}^{(0)},
\end{align*}
where
\begin{align*}
    \mathbf{\tilde{W}}^{x(1)}_{U,j,:}&= \Au{oc(1)}_{\lambda,j,:}\W{ox}+\Au{fc(1)}_{\lambda,j,:}\W{fx}+\Au{ig(1)}_{\lambda,j,:}\W{ix}\\
    &+\Au{ig(1)}_{\Delta,j,:}\W{gx},\\
    \mathbf{\tilde{W}}^{a(1)}_{U,j,:}&= \Au{oc(1)}_{\lambda,j,:}\W{oa}+\Au{fc(1)}_{\lambda,j,:}\W{fa}+\Au{ig(1)}_{\lambda,j,:}\W{ia}\\
    &+\Au{ig(1)}_{\Delta,j,:}\W{ga},\\
    \mathbf{\tilde{b}}^{(1)}_{U,j}&= [\Au{oc(1)}_{\lambda,j,:}\bias{o}+\Au{fc(1)}_{\lambda,j,:}\bias{f}+\Au{ig(1)}_{\lambda,j,:}\bias{i}+\Au{ig(1)}_{\Delta,j,:}\bias{g}\\
    &+\sum^s_{r=1}(\Au{oc(1)}_{\varphi,j,r}+\Au{fc(1)}_{\varphi,j,r}+\Au{ig(1)}_{\varphi,j,r})]+\bias{F}_j.
\end{align*}
A closed-form global upper bound $\gamma^U_j$ can thereafter be obtained by applying Holder's inequality:
\begin{align*}
    \gamma^U_j &= \epsilon\|\mathbf{\tilde{W}}^{x(1)}_{U,j,:}\|_q+\mathbf{\tilde{W}}^{x(1)}_{U,j,:}\x^{(1)}_0+\mathbf{\tilde{W}}^{a(1)}_{U,j,:}\mathbf{a}^{(0)}+\mathbf{\tilde{b}}^{(z)}_{U,j}\\
    &+\Au{fc(1)}_{\Delta,j,:}\mathbf{c}^{(0)}.
\end{align*}

An explicit function $F^L_j$ and a closed-form global lower bound $\gamma^L_j$ can also be found through similar steps such that $\gamma^L_j\leq F^L_j(\mathbf{X})\leq F_j(\mathbf{X})$, $\forall~\mathbf{X}\in\mathbb{R}^{n\times 2}$ where $\x^{(k)}\in\mathbb{B}_p(\x_0^{(k)},\epsilon)$. The derivations for a GRU follow similarly, and are put into the appendix (Section A.7) due to space constraint.

\subsection{Robustness Quantification Algorithm}
\label{sec:robust_cert}

As summarized in Table~\ref{tbl:bounds}, given a trained vanilla RNN, LSTM or GRU, input sequence $\mathbf{X}_0\in\mathbb{R}^{n\times m}$, $l_p$ ball parameters $p\geq 1$ and $\epsilon\ge 0$, for $\forall~j\in[t]$, $1/q=1-1/p$, there exist two fixed values $\gamma^L_j$ and $\gamma^U_j$ such that $\forall~\mathbf{X}\in\mathbb{R}^{n\times m}$ where $\x^{(k)}\in\mathbb{B}_p(\x_0^{(k)},\epsilon)$, the inequality $\gamma^L_j\leq F_j(\mathbf{X})\leq \gamma^U_j$ holds true. Now suppose the label of the input sequence is $j$, we aim at utilizing the uniform global bounds in Table~\ref{tbl:bounds} to find the largest possible lower bound $\epsilon_j$ of untargeted attacks. We formalize the objective and constraints as follows:
\begin{align*}
    \epsilon_j =\max_{\epsilon}\epsilon,\;
    \textrm{\textrm{s.t.}}\; \gamma^L_j(\epsilon ) \geq \gamma^U_i(\epsilon) ,\; \forall i\neq j.
\end{align*}
To verify the largest possible lower bound $\hat \epsilon$ of targeted attacks (target class be $i$), we solve the following:
\begin{align*}
    \hat \epsilon(i,j) =\max_{\epsilon}\epsilon,\;
    \textrm{\textrm{s.t.}}\; \gamma^L_j(\epsilon ) \geq \gamma^U_i(\epsilon).
\end{align*}
One can then verify that $\epsilon_j=\min_{i\neq j}\hat \epsilon(i,j)$. Recalling from the derivations in Sections~\ref{Exm:1} and~\ref{Exm:2}, slopes and intercepts utilized are functions of the input ranges. Therefore, the global lower bounds $\gamma^L$ and upper bounds $\gamma^U$ are also implicitly dependent on $\epsilon$, preventing the above optimization problems from having an analytic solution. In this work, we conduct binary search procedures to compute the largest possible $\epsilon_{j}$ ($\hat \epsilon$). In practice, we start by initializing the $\epsilon$ to be $\epsilon_0> 0$ and deriving upper and lower bounds according to Table~\ref{tbl:bounds}. These are followed by a condition check that governs the next step. If  $\exists\; i\neq j$ such that $\gamma^U_i\geq \gamma^L_j$, then we decrease $\epsilon$; otherwise, we increase $\epsilon$. We repeat the above procedure until a predefined maximum iterations are reached or when convergence is achieved.\footnote{The binary search converges when the interval falls below the given tolerance.} Proposed algorithms are summarized in Section A.2 and A.5 in the appendix.

\begin{figure}[t]
    \centering
    \includegraphics[width=0.43\textwidth]{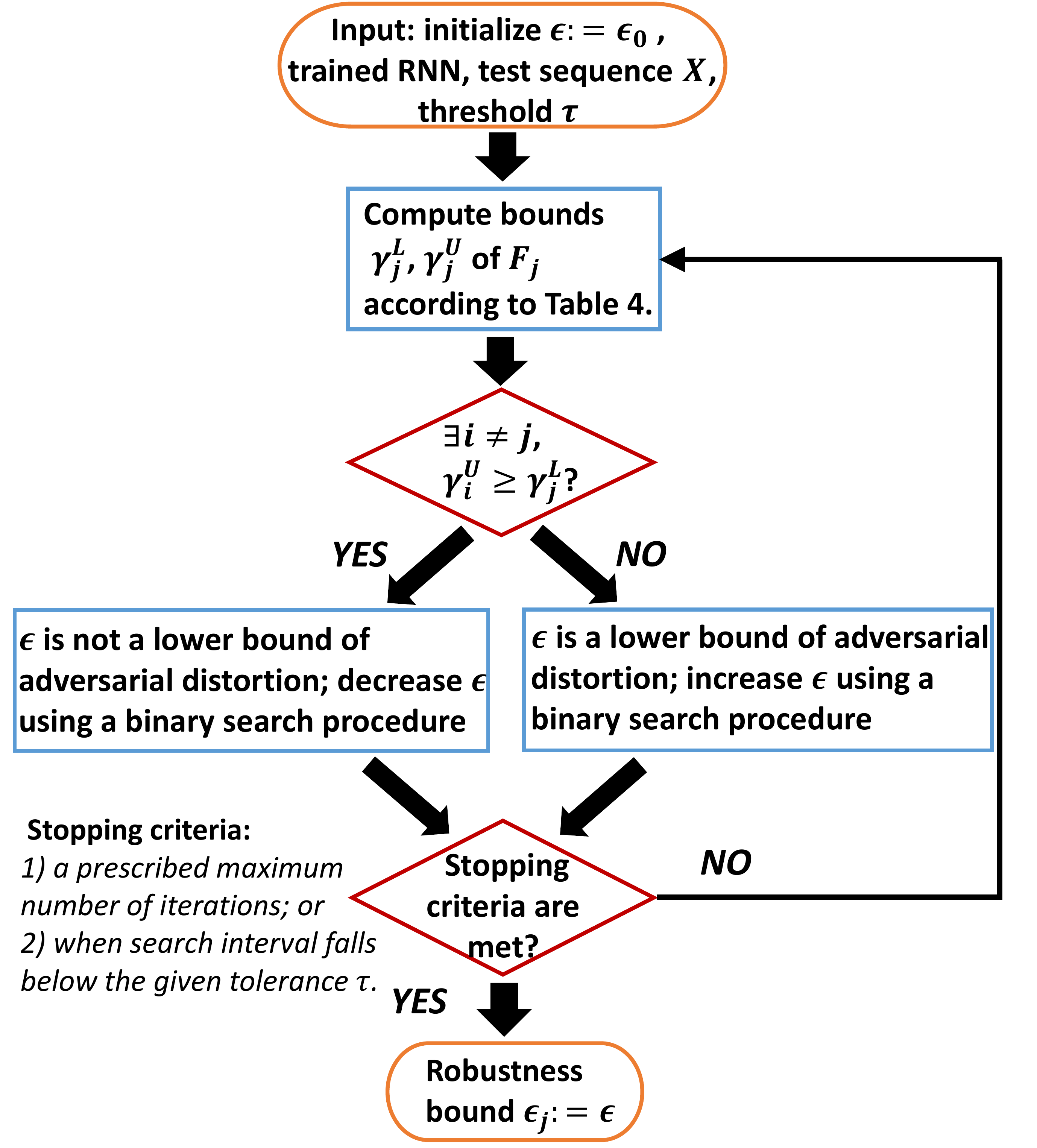}
    \caption{Steps in computing bounds for RNN networks.}
    \label{fig:flowchart}
    \vspace{-1.2em}
\end{figure}

\begin{table*}[tbh!]
\vspace{-0.5em}
\centering
\caption{(Experiment I) Averaged bounds and standard deviations ($\cdot$/$\cdot$) of POPCQRN and other baselines on MNIST dataset. POPQORN is the proposed method, CLEVER-RNN and C\&W-Ada are adapted from~\cite{weng2018evaluating} and~\cite{carlini2017towards}, respectively.} 
\label{tbl:exp1}
\scalebox{0.7}
{
\begin{tabular}{c|cccc||c|cccc}
\hline  \multirow{2}{*}{Network} & \multirow{2}{*}{$l_p$ norm} & Certified & Uncertified & Attack & \multirow{2}{*}{Network} & \multirow{2}{*}{$l_p$ norm} & Certified & Uncertified & Attack\\
 &  & (POPQORN) & (CLEVER-RNN) & (C\&W-Ada) &  &  & (POPQORN) & (CLEVER-RNN) & (C\&W-Ada)\\ \hline
RNN & $l_{\infty}$ & $0.0190/ 0.0047$ & $0.0831/ 0.0398$ & $0.2561/ 0.1372$ & RNN & $l_{\infty}$ & $0.0087/ 0.0018$ & $0.1259/ 0.0594$ & $0.3267/ 0.1804$ \\
4 layers & $l_2$ & $0.2026/ 0.0680$ & $0.8860/ 0.3920$ & $1.3768/ 0.6076$ & 14 layers & $l_2$ & $0.0526/ 0.0119$ & $0.6864/ 0.3172$ & $0.9278/ 0.4218$\\
32 hidden nodes & $l_1$ & $1.0551/ 0.2689$ & $5.0033/ 2.6034$ & $5.1592/ 2.4726$ & 64 hidden nodes & $l_1$ & $0.1578/ 0.0328$ & $2.6018/ 1.3142$ & $2.1149/ 0.9671$ \\\hline
RNN & $l_{\infty}$ & $0.0219/ 0.0047$ & $0.1099/ 0.0503$ & $0.2957/ 0.1374$ & LSTM & $l_{\infty}$ & $0.0202/ 0.0055$ & $0.0765/ 0.0356$ & $0.2546/ 0.1202$ \\
4 layers & $l_2$ & $0.2487/ 0.0592$ & $1.2006/ 0.5240$ & $1.7072/ 0.6757$ & 4 layers & $l_2$ & $0.2321/ 0.0636$ & $0.8302/ 0.3714$ & $1.3321/ 0.5529$ \\
64 hidden nodes & $l_1$ & $1.3199/ 0.2984$ & $6.4128/ 3.0360$ & $6.9347/ 3.1543$ & 32 hidden nodes & $l_1$ & $1.1913/ 0.3385$ & $4.6858/ 2.2761$ & $4.9689/ 2.2740$ \\\hline
RNN & $l_{\infty}$ & $0.0243/ 0.0050$ & $0.1405/ 0.0638$ & $0.3537/ 0.1469$ & LSTM & $l_{\infty}$ & $0.0218/ 0.0052$ & $0.1032/ 0.0470$ & $0.2893/ 0.1260$ \\
4 layers & $l_2$ & $0.2818/ 0.0557$ & $1.4698/ 0.6595$ & $2.1666/ 0.8556$ & 4 layers & $l_2$ & $0.2448/ 0.0618$ & $1.0947/ 0.4403$ & $1.6888/ 0.6705$ \\
128 hidden nodes & $l_1$ & $1.4362/ 0.2880$ & $7.7504/ 3.6968$ & $8.8906/ 4.2051$ & 64 hidden nodes & $l_1$ & $1.2644/ 0.3250$ & $6.0216/ 2.8502$ & $6.6931/ 2.9917$ \\\hline
RNN & $l_{\infty}$ & $0.0131/ 0.0031$ & $0.0841/ 0.0421$ & $0.2424/ 0.1291$ & LSTM & $l_{\infty}$ & $0.0218/ 0.0044$ & $0.1227/ 0.0508$ & $0.3303/ 0.1358$\\
7 layers & $l_2$ & $0.1045/ 0.0371$ & $0.6846/ 0.3283$ & $1.0300/ 0.4527$ & 4 layers & $l_2$ & $0.2491/ 0.0523$ & $1.3225/ 0.5412$ & $1.9379/ 0.7593$ \\
32 hidden nodes & $l_1$ & $0.4492/ 0.1088$ & $3.2046/ 1.7024$ & $2.9425/ 1.4266$ & 128 hidden nodes & $l_1$ & $1.2839/ 0.2760$ & $6.8665/ 3.0725$ & $7.7319/ 3.2369$ \\\hline
RNN & $l_{\infty}$ & $0.0131/ 0.0028$ & $0.1013/ 0.0525$ & $0.2660/ 0.1330$ & LSTM & $l_{\infty}$ & $0.0165/ 0.0044$ & $0.0924/ 0.0446$ & $0.2635/ 0.1290$ \\
7 layers & $l_2$ & $0.1113/ 0.0246$ & $0.8121/ 0.4210$ & $1.1454/ 0.4931$ & 7 layers & $l_2$ & $0.1400/ 0.0369$ & $0.7287/ 0.3434$ & $1.0733/ 0.4694$\\
64 hidden nodes & $l_1$ & $0.4499/ 0.0951$ & $3.5476/ 1.8650$ & $3.4262/ 1.6191$ & 32 hidden nodes & $l_1$ & $0.5680/ 0.1563$ & $3.2266/ 1.7095$ & $3.2309/ 1.4733$ \\\hline
RNN & $l_{\infty}$ & $0.0083/ 0.0023$ & $0.0931/ 0.0506$ & $0.2891/ 0.2004$ & LSTM & $l_{\infty}$ & $0.0099/ 0.0028$ & $0.1220/ 0.0676$ & $0.3148/ 0.1700$ \\
14 layers & $l_2$ & $0.0493/ 0.0139$ & $0.5074/ 0.2538$ & $0.7577/ 0.3859$ & 14 layers & $l_2$ & $0.0593/ 0.0148$ & $0.6688/ 0.3346$ & $0.9088/ 0.3854$\\
32 hidden nodes & $l_1$ & $0.1459/ 0.0399$ & $1.9323/ 1.0724$ & $1.6339/ 0.8493$ & 32 hidden nodes & $l_1$ & $0.1805/ 0.0496$ & $2.4880/ 1.3743$ & $2.0035/ 0.9447$ \\\hline
\end{tabular}
}
\vspace{-1em}
\end{table*}

\section{Experiments}
\label{sec:Exp}
\vspace{-0.2em}
\paragraph{Methods.} The proposed POPQORN is used to quantify robustness for all models herein. 
To the best of our knowledge, there is \textbf{no} previous work done on quantifying robustness with guarantees for RNNs. Therefore, we compare our results with CLEVER score~\cite{weng2018evaluating} (an estimation of the minimum adversarial distortion) and C\&W attack~\cite{carlini2017towards} (an upper bound of the minimum adversarial distortion). We emphasize on analyzing the characteristics of the certified bounds obtained by POPQORN and new insights they lead to. 
For comparison, we adapt CLEVER score to accommodate sequential input structure and denote it as CLEVER-RNN. Specifically, let $j$ and $i$ be the true and target labels, respectively. Assume \mbox{$g(\mathbf{X})=F_j(\mathbf{X})-F_i(\mathbf{X})$} is a Lipschitz function, and define \mbox{$L^t_q=\max{\{\|\nabla_t g(\mathbf{X})\|_q: \x^{(k)}\in\mathbb{B}_p(\x_0^{(k)},\epsilon_0),\forall k\in[m]\}}$} where $\nabla_t g(\mathbf{X})=(\frac{\partial g(\mathbf{X})}{\partial \x^{(t)}_1},\ldots,\frac{\partial g(\mathbf{X})}{\partial \x^{(t)}_n})^T.$ CLEVER-RNN score is then given as $\min{\{\frac{g(\mathbf{X}_0)}{\sum^m_{t=1}L^t_q},\epsilon_0\}}$\footnote{CLEVER score is defined by \mbox{$\min{\{\frac{g(\mathbf{X}_0)}{L_q},\epsilon_0\}}$}, where \mbox{$L_q=\max{\{\|(\nabla_1 g(\mathbf{X}),\ldots,\nabla_m g(\mathbf{X}))\|_q: \mathbf{X}\in\mathbb{B}_p(\mathbf{X}_0,\epsilon_0)\}}$}. We refer readers to the appendix Sec. B.2 for the details.}. We also adapt C\&W attack for our task and denote it as C\&W-Ada\footnote{We refer readers to the appendix Sec. B.3 for the details.}. The adapted C\&W-Ada puts higher weights on finding a successful attack, and prioritizes the minimization of distortion magnitude when an attack is found. We use the maximum perturbation of all frames as the C\&W score. We implement our algorithm using PyTorch to enable the use of GPUs. Using a server consisting of 16 NVIDIA Tesla K80 GPUs, it took about $4$ hours to calculate certified bounds for $1000$ samples in LSTMs with $4$ frames, and one day in LSTMs with $14$ frames. Except quantifying for LSTMs, the remaining experiments could be finished in about $10$ hours using a single NVIDIA TITAN X GPU. More experimental details are given in the appendix Section B.

\vspace{-0.5em}
\paragraph{Experiment I.} 
In this experiment, we evaluate POPQORN and other baselines on totally $12$ vanilla RNNs and LSTMs trained on the MNIST dataset. 
Table~\ref{tbl:exp1} gives the certified lower bounds of models found by POPQORN, together with uncertified CLEVER-RNN scores, and upper bounds found by C\&W-Ada attacks. 
Bounds obtained by POPQORN, CLEVER-RNN, C\&W-Ada increase with numbers of hidden neurons $s$ in RNNs and LSTMs, and generally decrease as the number of network layers $m$ grows\footnote{The provable safety region ($l_p$ balls) provided by POPQORN are distributed across framelets. The overall distortion allowed for an input sample is computed by $m^{1/p}\epsilon$.}. The uncertified bounds computed by CLEVER-RNN are similar to those found by C\&W-Ada attacks, yet it should be noted that these bounds are without guarantees. A supplementary comparison among bounding techniques (2D bounding planes, 1D bounding lines, constants) for LSTM evaluations is provided in Appendix Section B.1 (accompanied with theorems in Section A.6).
\begin{figure}
    \centering
    \subfigure[digit ``$1$"]{
    \label{fig:exp2dg1}
    \includegraphics[width=0.49\linewidth]{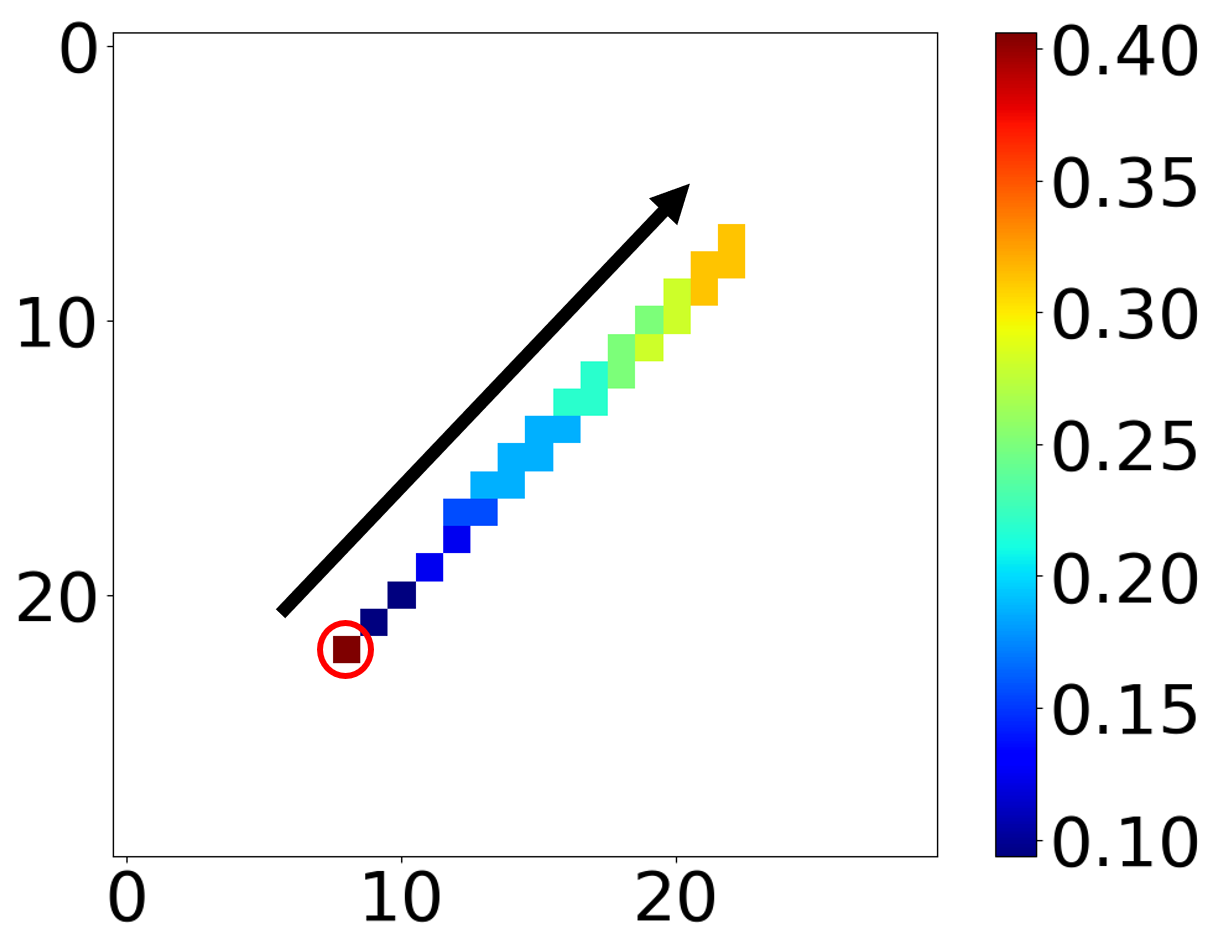}}\vspace{5pt}
    \subfigure[digit ``$4$"]{
    \label{fig:exp2dg4}
    \includegraphics[width=0.475\linewidth]{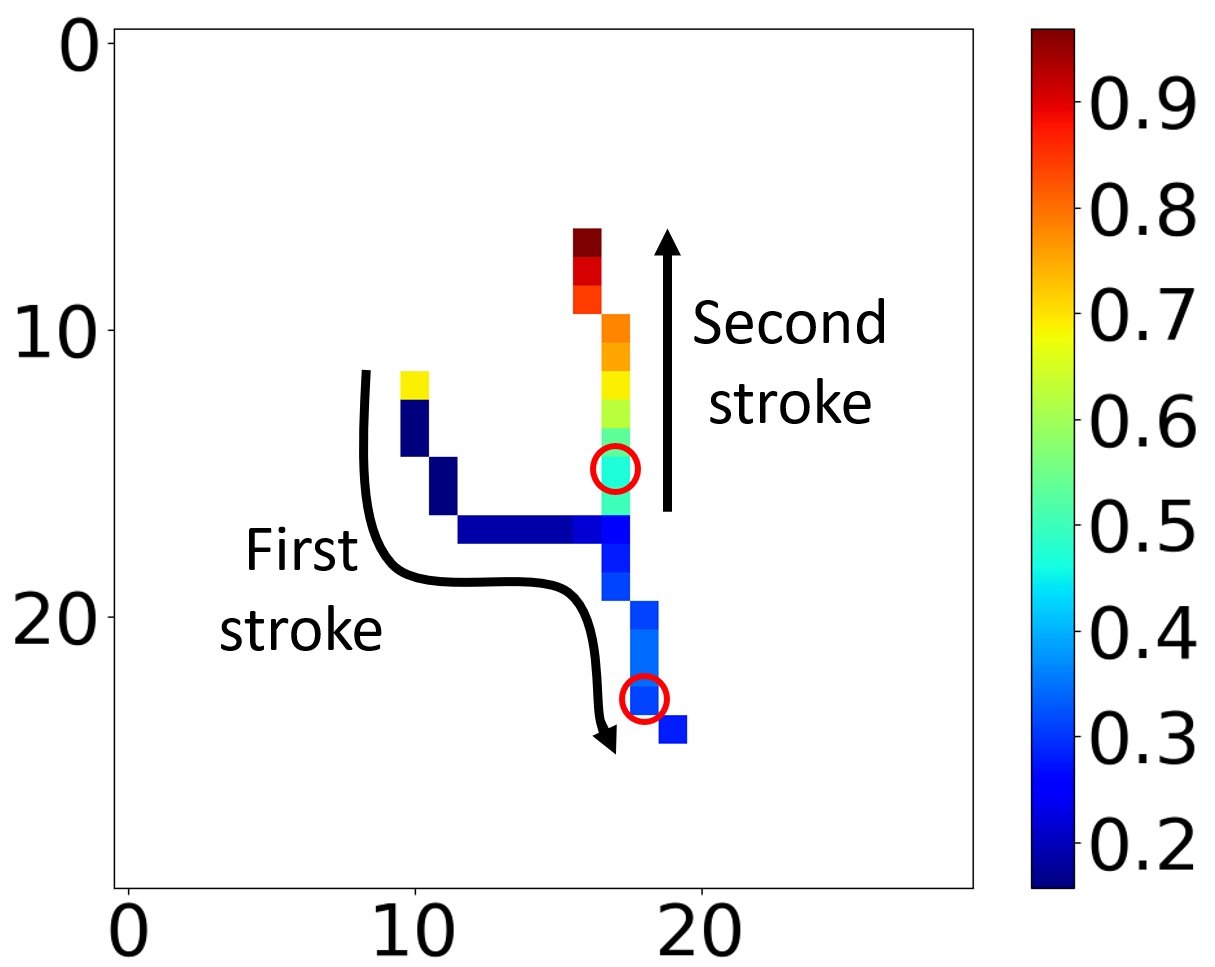}}
    \vspace{-1em}
    \caption{(Experiment II) Heat maps of certified $2$-norm bounds computed by POPQORN. Left (a): an example of digit ``$1$". Right (b): an example of digit ``$4$". Arrows indicate the order of strokes. }
    \label{fig:exp2}
    \vspace{-1.5em}
\end{figure}

\vspace{-0.5em}
\paragraph{Experiment II.} Next, we evaluate the proposed POPQORN quantification on LSTMs trained on the MNIST sequence dataset\footnote{Freely available at https://edwin-de-jong.github.io/blog/mnist-sequence-data/}.  Specifically, we compute the POPQORN bound on only one single input frame (i.e. we fix all input frames but one and derive certified bound for distortions on that frame ). After calculating bounds of all input frames, we identify the frames with minimal bounds and call them \textit{sensitive strokes}. In both subfigures of Figure~\ref{fig:exp2}, each point records the relative displacement from the previous point. Heat maps are used to track the changes in sensitivities (quantified robustness scores) with strokes . We identify the starting point in Figure~\ref{fig:exp2dg1} with a circle. Notably, this point has a relatively big certified bound. This implies the loose connection between the starting point of one's stroke and the number to be written down. Another tendency we observe is that points in the back have larger bounds compared with points in the front. Since the position of a point only affects the points behind it, points in the back have less influence on the overall shape. Therefore, they can tolerate perturbations of larger magnitude without influencing its classification result. In Figure~\ref{fig:exp2dg4}, we circle two points that are near the end of the first stroke and the start of the second stroke, respectively. These points have more influence on the overall shape of the sequence, which is also supported by the comparatively small POPQORN bounds.

\begin{figure}[t]
    \centering
    \includegraphics[width=0.48\textwidth]{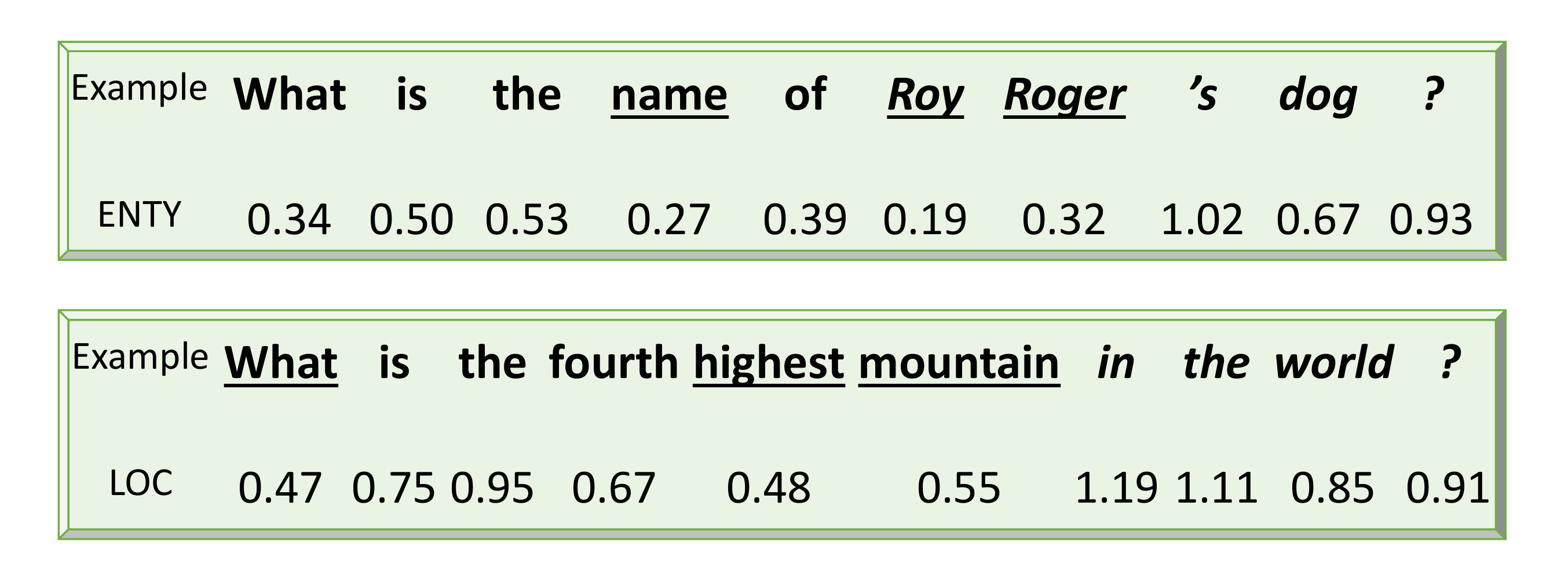}
    \vspace{-1.5em}
    \caption{(Experiment III) Two examples in the question classification task. The upper row gives the sample sentence; the lower row shows the POPQORN ($2$-norm) lower bounds of individual words. ``ENTY" and ``LOC" represent \textit{entity} and \textit{location}, respectively.}
    \label{fig:exp3}
    \vspace{-1.5em}
\end{figure}

\vspace{-1em}
\paragraph{Experiment III:} Lastly, POPQORN is evaluated on LSTM classifiers for the question classification (UIUC's CogComp QC) Dataset~\cite{li2002learning}\footnote{Freely available at http://cogcomp.org/Data/QA/QC/}. Figure~\ref{fig:exp3} shows two sample sentences in the UIUC's CogComp QC dataset. We conduct POPQORN quantification on individual steps (words), which guarantee robust classifier decisions as long as the perturbation magnitude in the word embedding space is smaller than the certificate. The 3 most sensitive words (words with 3 smallest bounds) are underlined. In the first example, the question is classified as ``ENTY"(entity). Correspondingly, \textbf{name} is among the three most sensitive words, which is consistent with human cognition. In the second example, the question is classified as ``LOC"(location). Again, \textbf{mountain} is shortlisted in the top three sensitive words. More examples are provided in the appendix Section B.4. Such observed consistency suggests POPQORN's potential in distinguishing the importance of different words.

\section{Conclusion}
\vspace{-0.2em}
This paper has proposed, for the first time, a robustness quantification framework called \emph{POPQORN} that handles various RNN architectures including vanilla RNNs, LSTMs and GRUs. The certified bound gives a guaranteed lower bound of the minimum distortion in RNN adversaries, in contrast to the upper bound suggested by attacks. Experiments on different RNN tasks have demonstrated that POPQORN can compute non-trivial certified bounds, and provide insights on the importance and sensitivity of a single frame in sequence data.

\section*{Acknowledgment}
The authors would like to thank Zhuolun Leon He for useful discussion. 
This work is partially supported by the Big Data Collaboration Research grant from SenseTime Group (CUHK Agreement No. TS1610626), the General Research Fund (Project 14236516, 17246416) of the Hong Kong Research Grants Council, MIT-IBM program, MIT-Skoltech program, and MIT-SenseTime program.

\bibliography{ref}
\bibliographystyle{icml2019}

\end{document}